\documentclass[pmlr]{jmlr}

\RequirePackage{graphicx}
\usepackage{booktabs}
\usepackage{longtable}
\usepackage{multirow}
\usepackage{xcolor}
\usepackage{colortbl}
\usepackage{siunitx}
\usepackage{float}
\usepackage{microtype}
\usepackage{subcaption}
\usepackage{hyperref}
\usepackage{amsmath}
\usepackage{amssymb}
\usepackage{mathtools}
\usepackage{bm}

\jmlrproceedings{PMLR}{Proceedings of Machine Learning Research}
\jmlrvolume{340}
\jmlryear{2026}
\jmlrworkshop{Machine Learning for Healthcare}

\title[Negation-Aware Selective Training]{%
\shortstack[c]{%
NAST: Improving Negation Handling\\
in Medical Vision–Language Models\\
through Negation-Aware Selective Training}}

\author{
    \Name{Ali Abbasi} \Email{aiai@udel.edu}\\
    \addr University of Delaware
\AND
    \Name{Mehdi Taghipour} \Email{mtaghipour@usaclinics.com}\\
    \addr Cleveland Clinic; USA Vein Clinics
\AND
    \Name{Rahmatollah Beheshti} \Email{rbi@udel.edu}\\
    \addr University of Delaware
}

\begin{document}

\maketitle

\begin{abstract}
Negation is a fundamental linguistic operation in clinical reporting, yet vision--language models (VLMs) frequently fail to distinguish affirmative from negated medical statements. To systematically characterize this limitation, we introduce MedNeg-Bench, a radiology-specific diagnostic benchmark that evaluates polarity sensitivity under controlled clinical conditions, revealing that common medical VLMs consistently confuse negated and non-negated findings. To enable learning beyond simple condition absence, we further construct MedNeg-FT, a contextual clinical negation dataset that encodes structured claims and supports attribute-level negations involving location and severity. Building on these resources, we propose Negation-Aware Selective Training (NAST), an interpretability-guided adaptation method that uses causal tracing effects (CTEs) to modulate layer-wise gradient updates during fine-tuning. NAST scales each layer’s update according to its causal contribution to negation processing, transforming mechanistic interpretability signals into a principled optimization rule. Experiments demonstrate improved discrimination of affirmative and negated clinical statements without degrading general vision--language alignment, highlighting the value of causal interpretability for targeted model adaptation in safety-critical medical settings. Code and resources are available at \url{https://github.com/healthylaife/NAST}.
\end{abstract}

\begin{figure}[t]
    \centering
    \includegraphics[width=0.6\linewidth]{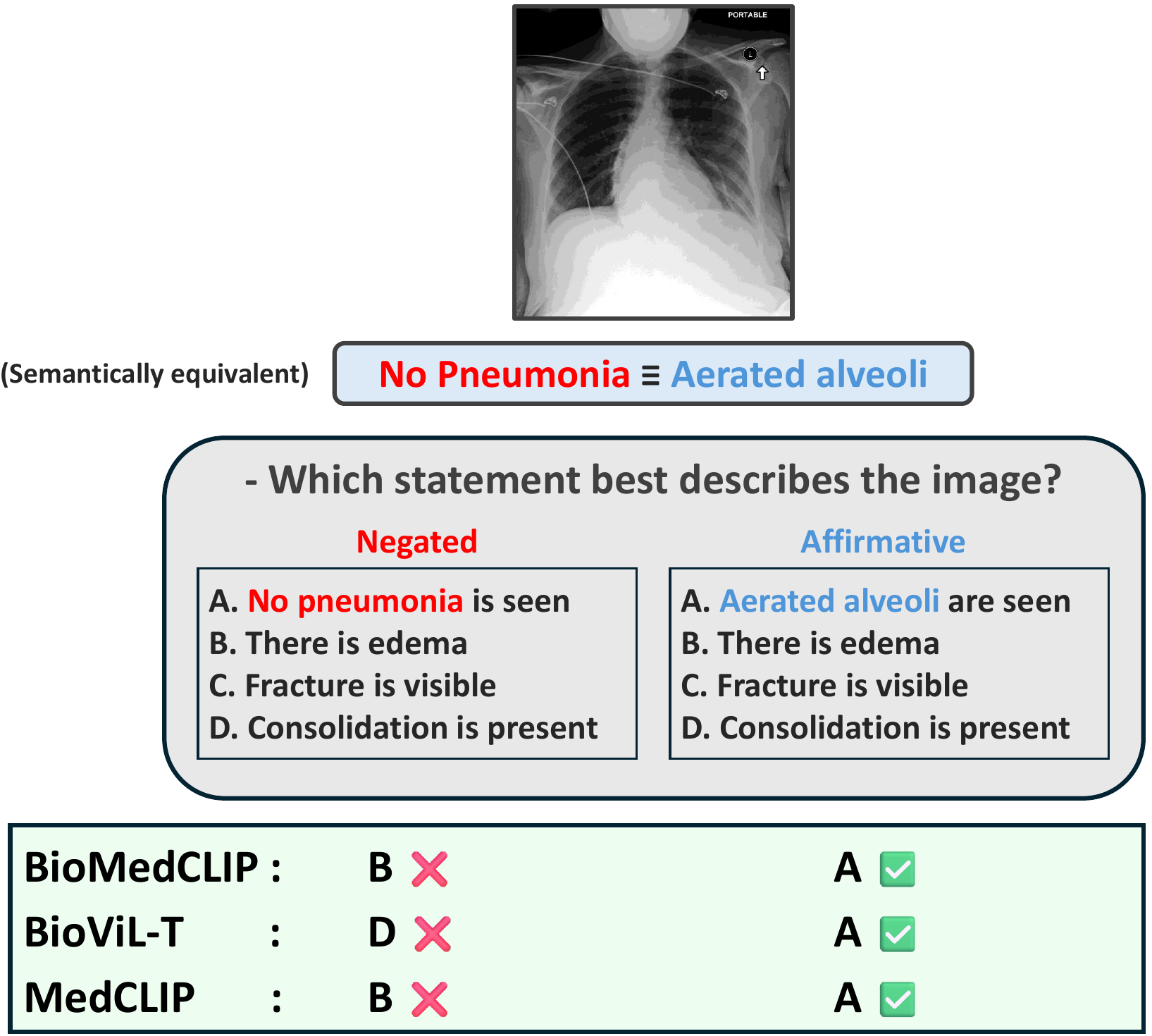}
    \caption{
    \textbf{Negation failure under polarity-controlled medical descriptions.}
    For the same chest X-ray, two multiple-choice sets are identical except for a single semantically equivalent phrase: a negated finding (``No pneumonia is seen'') versus its affirmative counterpart (``Aerated alveoli are seen''). Despite the minimal wording change and shared clinical meaning, representative medical VLMs predict correctly under the affirmative phrasing but fail under the negated phrasing, illustrating a systematic affirmative bias.
    }
    \label{fig:teaser_negation_gap}
\end{figure}

\section{Introduction}
Vision–language models (VLMs) have achieved impressive performance across a range of multimodal tasks, including image–text retrieval, visual question answering, and medical report understanding \citep{radford2021clip, li2022blip, alayrac2022flamingo}. In clinical settings, such models are increasingly used to associate radiological images with textual descriptions, enabling applications such as automated report generation and decision support \citep{wang2022medclip, zhang2023biomedclip, lin2023pmc, bannur2023learning}. However, despite strong aggregate performance, recent evidence suggests that VLMs often rely on shallow correlations between visual features and textual tokens, raising concerns about their robustness to linguistically subtle but clinically critical distinctions \citep{yuksekgonul2022and, liu2026seeing}.

One such distinction is a surprisingly simple linguistic phenomenon—negation \citep{vatsa2025no}. In radiology reports, negation reverses the clinical interpretation of visual findings. Statements such as “no pleural effusion” or “no consolidation in the right lower lobe” convey the absence of pathology and directly influence downstream clinical decisions. Confusing negated and affirmative statements can lead to serious misinterpretations, even when the underlying visual evidence is correctly perceived \citep{ko2025bringing}. While negation has been studied in natural language processing \citep{chapman2001simple, morante2012conandoyle, ettinger2020bert}, its interaction with multimodal representation learning—particularly in medical VLMs—remains poorly understood \citep{alhamoud2025vlmnegation}.

Negation is challenging for vision–language models because it requires reasoning about how linguistic modifiers alter visual meaning. Since contrastive pretraining is dominated by affirmative image–caption pairs, descriptions of absence (e.g., “no”, “without”) are relatively underrepresented \citep{ko2025bringing}. Although recent work has examined negation in multimodal models \citep{singh2024learn, cai2025tng, han2026negation, alhamoud2025vlmnegation}, most studies focus on general-domain object presence. In medical imaging, however, negation often targets specific attributes—such as location or severity—rather than existence alone (e.g., “No large pleural effusion” or “Consolidation not in the right lower lobe”). Consequently, negation understanding in medical VLMs remains insufficiently explored despite its clinical importance.

In this work, we first show that well-known medical vision–language models exhibit systematic failures in negation understanding. To isolate this behavior, we introduce MedNeg-Bench, a radiology-specific diagnostic benchmark that evaluates polarity sensitivity under tightly controlled clinical conditions. MedNeg-Bench constructs paired statements that are identical in structure and differ only in clinical polarity (e.g., “Cardiomegaly is seen” vs. “Normal heart size is seen”). Despite their semantic equivalence, models consistently perform better on affirmative formulations than on their negation-equivalent counterparts, revealing a consistent affirmative bias (Figure~\ref{fig:teaser_negation_gap}).

Addressing this limitation requires more than simply adding negated examples, as most prior approaches do \citep{singh2024learn, ko2025bringing, cai2025tng, han2026negation, alhamoud2025vlmnegation}. In clinical language, negation is often contextual and attribute-specific, targeting not only condition existence but also location or severity \citep{chapman2001simple, harkema2009context}. To model this granularity, we construct MedNeg-FT, a contextual clinical negation dataset based on structured radiological claims \citep{zhang2023cadchest}. Building on this dataset, we propose Negation-Aware Selective Training (NAST), an interpretability-guided adaptation method for CLIP-based medical VLMs. NAST uses causal tracing to estimate each layer’s contribution to negation processing and scales gradient updates accordingly, assigning larger updates to causally influential layers. Experiments show improved discrimination between affirmative and negated clinical statements without degrading general vision–language alignment. 

In particular, this paper makes the following contributions:

\begin{itemize}
    \item \textbf{Diagnostic evaluation of negation in medical VLMs:} We introduce MedNeg-Bench, a radiology-specific diagnostic benchmark that isolates polarity sensitivity by constructing tightly controlled affirmative–negated statement pairs, revealing systematic failures and an affirmative bias in well-known medical vision–language models.
    \item \textbf{Contextual clinical negation dataset:} We propose a new fine-tuning dataset based on structured radiological claims that captures attribute-level negation—including location and severity—enabling supervision that reflects how negation is expressed in real clinical reports beyond simple condition absence.
    \item \textbf{Negation-Aware Selective Training (NAST):} We present NAST, an interpretability-guided adaptation method for CLIP-based medical VLMs that uses causal tracing effects to modulate layer-wise gradient updates, allocating learning capacity according to each layer’s causal involvement in negation processing.
    \item \textbf{Empirical validation of interpretability-guided optimization:} We show that NAST improves discrimination between affirmative and negated clinical statements while preserving performance on standard multimodal tasks, demonstrating that mechanistic interpretability can guide effective model adaptation in medical settings.
\end{itemize}

\subsection{Generalizable Insights about Machine Learning in Healthcare}

Our findings highlight several generalizable insights for the development of machine learning systems in healthcare. First, correctly modeling negation is critical for clinical reliability, as many medical decisions depend on distinguishing the presence versus absence of findings (e.g., ruling out pneumothorax or confirming no evidence of edema). Small linguistic variations in polarity can correspond to clinically significant differences, and failure to capture these distinctions may lead to incorrect or unsafe interpretations.

Second, our results suggest that standard vision--language models exhibit systematic biases toward affirmative descriptions, even when trained on large-scale medical data. This indicates that scaling data alone is insufficient to address clinically important reasoning gaps, and that targeted training strategies are necessary to improve robustness in safety-critical settings.

Third, we show that fine-grained supervision based on structured clinical attributes (e.g., condition, location, severity) enables more precise control over model behavior. This highlights the importance of leveraging domain-specific structure and expert knowledge when designing training objectives for medical AI systems. Finally, our approach demonstrates that improving clinically relevant reasoning does not require full model retraining. Instead, selective and parameter-efficient adaptation can yield substantial gains, suggesting a practical pathway for updating deployed models in real-world healthcare environments where computational and regulatory constraints are significant.

\section{Related Work}

\paragraph{Negation in NLP and Multimodal Models}

Negation has long been recognized as a core challenge in natural language understanding, particularly in the clinical domain, where correctly interpreting negated findings is critical for safe decision-making. Early NLP work developed rule-based and contextual approaches to detect negated and uncertain medical concepts, demonstrating that negation often operates over specific attributes of a condition rather than indicating its absence \citep{chapman2001simple, harkema2009context, morante2012conandoyle}. Analyses of pretrained language models later revealed that, despite contextual representations, models still struggle with polarity reversals and compositional negation, frequently relying on surface-level lexical cues \citep{ettinger2020bert, hosseini2021understanding, vatsa2025no}.

More recently, negation has gained attention in multimodal and vision--language learning \citep{zhang2025negvqa}. Singh et al.\ introduced CC-Neg and showed that VLMs often fail to associate negated captions with images, motivating contrastive fine-tuning strategies \citep{singh2024learn}. Subsequent work proposed multi-level retrieval and multiple-choice diagnostics revealing persistent affirmation bias \citep{alhamoud2025vlmnegation}, training-time negation generation pipelines \citep{cai2025tng}, and test-time adaptation methods addressing distribution shifts between affirmative and negated concepts \citep{han2026negation}. Further efforts have pursued LLM-driven negation-inclusive data generation for CLIP fine-tuning \citep{park2025know}, structured reasoning and token-merging strategies \citep{kang2026not}, and zero-shot approaches that exploit compositional arithmetic in CLIP's embedding space \citep{aggarwal2026seeing}. In the medical domain, Ko and Park introduced a negation-aware benchmark and contrastive framework for chest X-ray interpretation \citep{ko2025bringing}.

Despite these advances, existing multimodal negation studies have largely focused on general-domain object absence or data-centric augmentation strategies. Comparatively little attention has been paid to the structured, attribute-level negation that characterizes clinical language, or to how negation is distributed across model layers and learning dynamics.

\paragraph{Medical Vision–Language Models}

Medical vision–language models adapt large-scale multimodal pretraining to the clinical domain by aligning medical images with structured or free-text reports. Early efforts focused on radiology, leveraging paired chest X-rays and reports to learn joint embeddings for retrieval, classification, and report generation. Notable examples include ConVIRT \citep{zhang2022contrastive} and GLoRIA \citep{huang2021gloria}, which demonstrated that contrastive pretraining on image–report pairs improves downstream diagnostic performance compared to vision-only models.

Building on CLIP-style architectures \citep{radford2021clip}, several works introduced domain-specific adaptations. MedCLIP \citep{wang2022medclip} replaces web-scale image–text data with curated medical image–report pairs and introduces medical-aware text encoders. BioMedCLIP \citep{zhang2023biomedclip}, PubMedCLIP \citep{eslami2023pubmedclip}, and PMC-CLIP \citep{lin2023pmc} further scale this paradigm by pretraining on large biomedical corpora, including PubMed Central figures and captions, yielding strong zero-shot performance across medical imaging tasks. More recent models, such as Med-Flamingo \citep{moor2023med}, UniMed-CLIP \citep{khattak2024unimed}, and QwenCLIP \citep{wei2025qwenclip}, explore instruction tuning and multimodal prompting to better align model outputs with clinical reasoning. However, systematic evaluation of negation handling in medical vision–language models remains limited, highlighting a gap that this work aims to address.

\section{MedNeg-Bench: A Benchmark for Negation in Medical VLMs}
\label{sec:medneg-bench}
Recent work has begun to benchmark negation understanding in vision–language models through tasks such as image retrieval and multiple-choice question answering \citep{singh2024learn, alhamoud2025vlmnegation}. While these benchmarks reveal performance gaps between captions containing negation and those without, they typically do not enable direct comparison between negated and affirmative descriptions that convey the same semantic state. As a result, it remains unclear whether observed failures reflect genuine difficulty in reasoning about negation, or broader weaknesses in compositional language–vision alignment.

The medical domain offers a unique opportunity to isolate this phenomenon. Unlike general-domain scenes—where the absence of an object (e.g., “no car”) does not admit a single affirmative equivalent—many negated clinical findings can be naturally expressed using affirmative alternatives without explicit negation. For example, the absence of pneumonia may be described either as “no pneumonia” or “aerated lungs.” This enables controlled evaluation using semantically equivalent sentence pairs differing only in polarity. Motivated by this observation and the lack of benchmarks targeting negation understanding in medical VLMs, we construct MedNeg-Bench, a diagnostic benchmark that directly contrasts negated and affirmative clinical descriptions under identical visual contexts.

\subsection{Benchmark Construction}
\label{sec:benchmark}

\textbf{Data sources and cohort.}
MedNeg-Bench is constructed from chest X-ray images and structured labels derived from CheXpert \citep{irvin2019chexpert}, which in turn is based on MIMIC-CXR-JPG \citep{johnson2024mimiccxrjpg}. These datasets consist of de-identified studies collected from hospital systems and are widely used for developing medical vision--language models. We focus on standard frontal chest X-ray studies from adult patients, following common practice in prior work.

CheXpert provides labels for 14 radiological conditions, annotated as present (1), absent (0), uncertain ($-1$), or not mentioned. To ensure consistency and clinical interpretability, we discard studies labeled \textit{No Finding} and retain those with at least two positive and three absent conditions. This filtering ensures that each constructed question includes multiple affirmed findings and at least one clinically meaningful negated condition. The resulting cohort contains 6{,}965 studies spanning common thoracic pathologies such as atelectasis, cardiomegaly, pleural effusion, and pulmonary edema.

\begin{figure*}[t]
    \centering
    \includegraphics[width=0.98\linewidth]{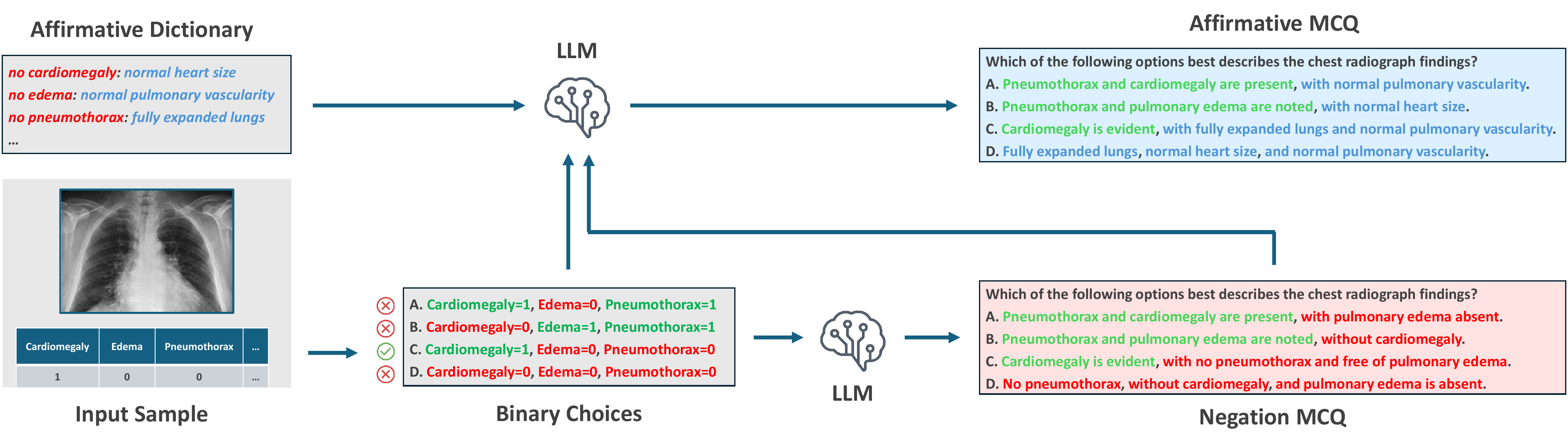}
    \caption{
        \textbf{MedNeg-Bench benchmark construction pipeline.}
        Starting from MIMIC-CXR images and binary diagnostic labels, the pipeline generates 
        (1) structured label permutations, (2) LLM-generated negation MCQs, 
        (3) affirmative rewrites, and (4) final paired MCQs. 
        All mappings and outputs were reviewed by two board-certified radiologists. 
    }
    \label{fig:benchmark_pipeline}
\end{figure*}

\textbf{Benchmark construction.}
To enable controlled polarity comparisons (i.e., paired descriptions differing only in clinical polarity), we curate affirmative alternatives for each absent condition (e.g., \textit{no cardiomegaly} $\leftrightarrow$ \textit{normal heart size}) in consultation with two board-certified radiologists (Appendix~\ref{app:affirmative}). For each study, we generate paired multiple-choice questions (MCQs) in three steps: (1) construct contrastive label configurations by permuting condition presence and absence to form hard negatives; (2) use an LLM to generate a radiology-style question with explicit negation; and (3) use a second LLM to produce an affirmative counterpart by replacing negated phrases with curated alternatives while preserving structure.

This yields MCQ pairs differing only in polarity, enabling direct comparison between negated and affirmative formulations. Data are split at the patient level to prevent leakage. Figure~\ref{fig:benchmark_pipeline} illustrates the pipeline; full details appear in Appendix~\ref{app:full}.

\subsection{Empirical Analysis on Medical VLMs}
\label{sec:mednegbench-analysis}

\begin{figure}[t]
    \centering
    \includegraphics[width=0.9\linewidth]{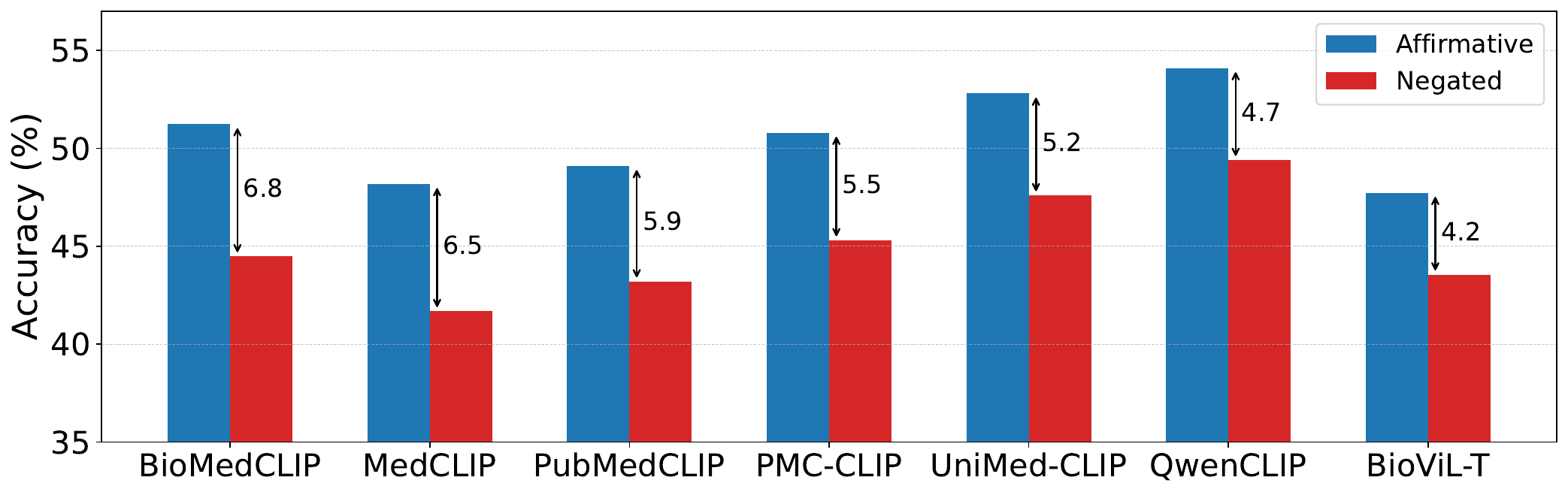}
    \caption{
    \textbf{Evaluation of medical vision--language models on MedNeg-Bench.} Despite identical sentence structure and equivalent underlying clinical meaning, all models exhibit consistently lower performance on negated formulations, revealing a systematic affirmative bias in medical VLMs. The gap between affirmative and negated accuracy highlights the extent and robustness of this bias across architectures.
    }
    \label{fig:med_vlm_negation_gap}
\end{figure}

We evaluate a diverse set of medical vision--language models, including BioMedCLIP \citep{zhang2023biomedclip}, MedCLIP \citep{wang2022medclip}, PubMedCLIP \citep{eslami2023pubmedclip}, PMC-CLIP \citep{lin2023pmc}, UniMed-CLIP \citep{khattak2024unimed}, QwenCLIP \citep{wei2025qwenclip}, and BioViL-T \citep{bannur2023learning}, on MedNeg-Bench using multiple-choice accuracy. All models are used with their publicly released pretrained checkpoints in a zero-shot setting, without additional fine-tuning. For each image--caption pair, we compute cosine similarity between the normalized image and text embeddings. Predictions are obtained by selecting the highest-scoring candidate within each multiple-choice set. Where applicable, we use the default logit scaling (temperature) provided by each model; no additional calibration is applied. All models are evaluated in full precision without quantization. Results are reported separately for affirmative and negated MCQ sets (Figure~\ref{fig:med_vlm_negation_gap}).

Despite identical structure and clinical meaning across affirmative and negated MCQ sets, all models exhibit a consistent performance gap favoring affirmative formulations. This gap persists despite the known difficulty of contrastive models in handling descriptive adjectives and attribute-level semantics (e.g., \textit{normal}, \textit{clear}, \textit{smooth}, \textit{sharp}) \citep{yuksekgonul2022and, buettner2024investigating}. Even though affirmative questions mostly rely on such adjective-based alternatives (Appendix~\ref{app:affirmative}), models still perform worse on negated formulations, suggesting that the degradation reflects a deeper deficiency in negation sensitivity rather than adjective understanding alone.

\subsection{Label Reliability and Benchmark Robustness}

A potential concern is that CheXpert labels, derived from automated report parsing, may contain noise, especially for negation. We mitigate this in both benchmark design and validation. First, our benchmark does not rely on detecting negation from free-text reports. Instead, we use structured condition labels (present vs.\ absent) and introduce negation deterministically via curated mappings (Appendix~\ref{app:affirmative}), ensuring controlled semantics independent of negation extraction errors.

Second, evaluation uses a contrastive multiple-choice setup with hard negatives, which reduces sensitivity to label noise compared to direct classification. All mappings and samples were also reviewed by two board-certified radiologists. Finally, we evaluate on the expert-annotated CheXpert-500 test set \citep{irvin2019chexpert} (Appendix~\ref{res:chexpert500}), where consistent improvements confirm that gains are not driven by artifacts of automated labeling.

\section{Method}
\label{sec:method}

The results above reveal that medical vision–language models struggle to reliably interpret negated clinical statements, motivating targeted adaptation. We therefore introduce \textbf{Negation-Aware Selective Training (NAST)}, a fine-tuning framework for CLIP-based medical VLMs. NAST combines a contextual negation dataset with causal tracing \citep{meng2022locating} to identify layers most influential for negation processing and selectively modulates their updates during training. By grounding optimization in layer-wise causal relevance, NAST improves negation sensitivity while preserving vision–language alignment.

\subsection{MedNeg-FT: A Contextual Negation Fine-Tuning Dataset}
\label{sec:dataset}

We construct MedNeg-FT, a large-scale contextual negation fine-tuning dataset from MIMIC-CXR \citep{johnson2024mimiccxrjpg} using CAD annotations \citep{zhang2023cadchest}, derived from radiology reports, which include condition, severity, location, and diagnostic uncertainty. 

\paragraph{Structured Clinical Facts.}
Each study provides disease-level annotations specifying condition, existence, location, and severity. We convert these into canonical structured facts,
\[
(\text{condition}, \text{existence}, \text{location}, \text{severity}),
\]
retaining only definitive labels and excluding uncertain cases.

\begin{figure*}[t]
    \centering
    \includegraphics[width=0.9\textwidth]{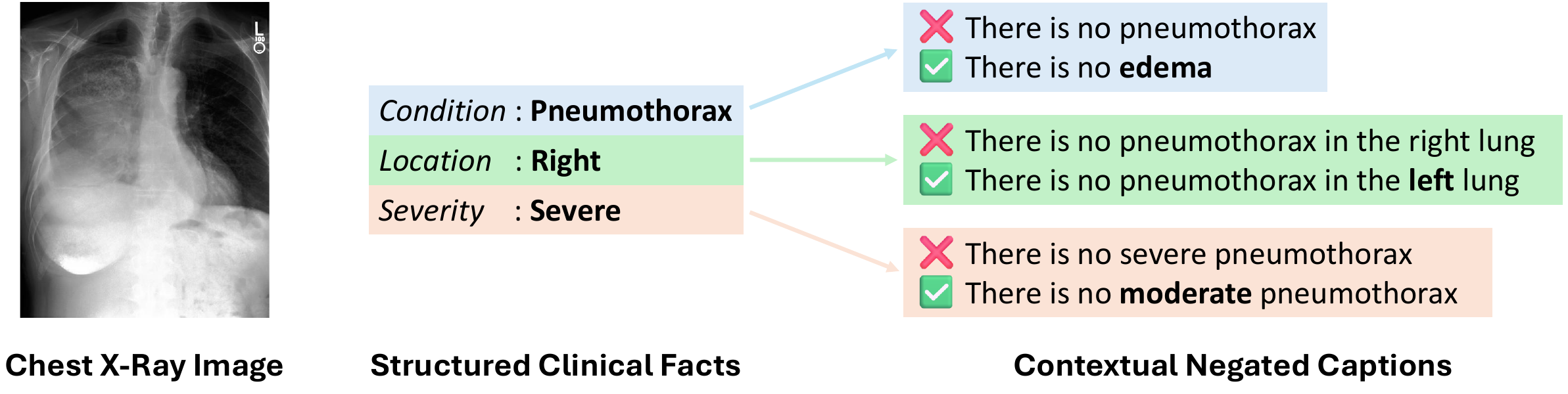}
    \caption{
    \textbf{Contextual negation in MedNeg-FT.}
    Each chest X-ray is associated with structured clinical facts describing condition, location, and severity. This information enables generation of negated captions that require models to reason over multiple attributes, rather than simple presence or absence.
    }
    \label{fig:contextual_negation}
\end{figure*}

\paragraph{Contextual Negation Construction.}
From each true fact, we generate contextual negations by minimally altering exactly one attribute while keeping others fixed. We consider three types: existence (present vs.\ absent), location (e.g., left vs.\ right), and severity (e.g., mild vs.\ severe). These controlled perturbations yield clinically plausible, fine-grained counterfactuals. Examples of such contextual negations are illustrated in Figure~\ref{fig:contextual_negation}.

\paragraph{Caption Generation and Formats.}
All facts are verbalized into radiology-style statements. We create two supervision formats:  
(i) \emph{claim-based contrast sets} with one true claim and multiple hard negatives, and  
(ii) \emph{single negated captions} for auxiliary contrast training.  

This process produces approximately one million image–text pairs, which we partition into training, validation, and test splits at the patient level to prevent data leakage. Compared to open-domain negation datasets \citep{alhamoud2025vlmnegation, singh2024learn}, our scale is sufficient due to the restricted ontology, precise annotations, and targeted counterfactual supervision. Additional details are provided in Appendix~\ref{app:fine}.

\subsection{Estimating Layer-Wise Causal Contribution to Negation}
\label{sec:causal-tracing}

Our framework derives layer-wise importance weights to modulate gradient updates during fine-tuning, targeting layers based on their contribution to negation processing. This design minimizes changes to less relevant layers, preserving pretrained capabilities. We adopt a causal tracing approach \citep{meng2022locating}, inspired by \citep{quantmeyer2024and}, and adapt it to the medical imaging domain and our training objective.

\paragraph{Causal Probing Data.}
We construct a causal probing set from CAD annotations \citep{zhang2023cadchest} with high certainty and severe disease labels. For each condition, we generate paired captions: a \emph{correct} caption describing the true clinical state (presence or absence) and a corresponding \emph{foil} caption describing the opposite state, constrained to have identical token lengths. Unlike prior work that uses vague quantifiers such as ``some'' \citep{quantmeyer2024and}, we introduce the modifier ``severe'' for affirmative findings (e.g., ``severe edema'' vs.\ ``no edema''), yielding a clearer semantic contrast in the medical setting.

\paragraph{Causal Tracing in CLIP.}
Given an image--caption pair, we compute CLIP similarity scores for the correct and foil captions, denoted $S^{\mathrm{corr}}$ and $S^{\mathrm{foil}}$, and define $d = S^{\mathrm{corr}} - S^{\mathrm{foil}}$ as the baseline negation sensitivity.

We record text-encoder hidden states from the foil forward pass. In a second pass with the correct caption, we intervene by replacing the hidden state at layer $\ell$ and token position $p$ with the corresponding foil representation, producing an intervened score $S^{\ell,p}$. Following \citep{quantmeyer2024and}, we compute
\[
d^{\ell,p} = S^{\mathrm{corr}} - S^{\ell,p}, 
\qquad
\mathrm{CTE}(\ell,p) = \frac{d^{\ell,p}}{d}.
\]
$\mathrm{CTE}(\ell,p)$ quantifies the fraction of negation-sensitive signal attributable to position $(\ell,p)$. CTE scores are computed once on this fixed probing set prior to fine-tuning and kept frozen throughout training.

\begin{figure}[t]
    \centering
    \includegraphics[width=0.6\linewidth]{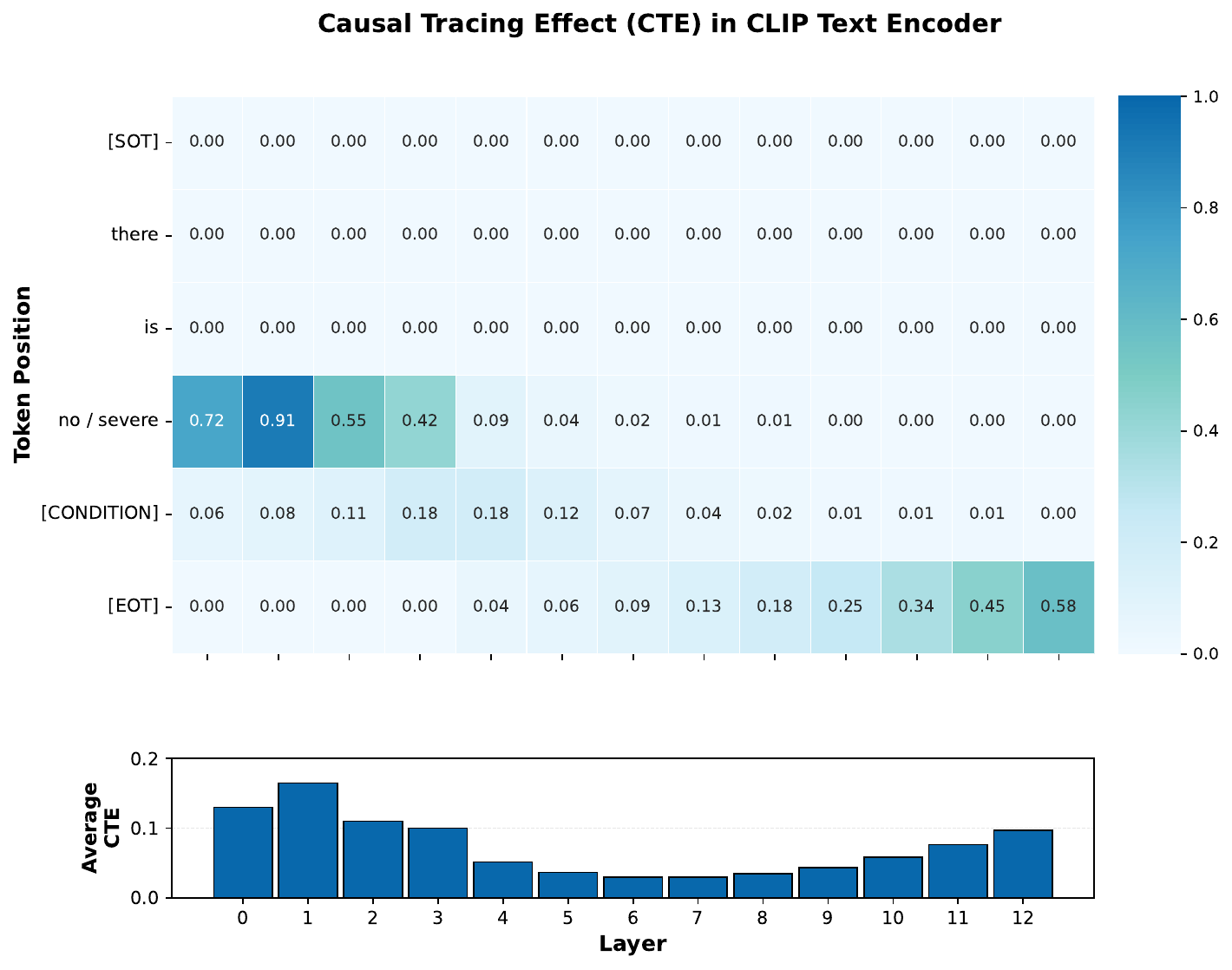}
    \caption{
    \textbf{Causal tracing effect (CTE) across layers and token positions in the CLIP text encoder.}
    The heatmap shows the average CTE per layer–position pair, merged across instances where negation appears in the caption and in the foil. The bar chart shows the layer-wise average CTE across all token positions. Negation-sensitive causal signal concentrates in early layers (0–3), then drops sharply at layer 4 and gradually recovers in later layers, driven primarily by signal accumulating at the [EOT] position.
    }
    \label{fig:cte_per_layers_and_positions}
\end{figure}

As shown in Figure~\ref{fig:cte_per_layers_and_positions}, the negation-sensitive signal is concentrated in early layers (0--3), rather than uniformly distributed, and partially re-emerges in later layers as it is consolidated at the \texttt{[EOT]} token. This indicates that negation processing is localized to specific stages of the CLIP text encoder rather than spread uniformly, with early layers performing the bulk of negation-relevant computation and later layers aggregating this signal into the final sequence representation. Additional details are provided in Appendix~\ref{app:construction}.

\subsection{CTE-Weighted Layer-Specific Adaptation}
\label{method:adaptaion}
We leverage the layer-wise causal tracing effects (CTEs) estimated in the previous subsection to guide selective fine-tuning of CLIP-based medical VLMs. To achieve parameter-efficient adaptation, we fine-tune the model using Low-Rank Adaptation (LoRA) \citep{hu2022lora} modules inserted into the transformer layers, while keeping the backbone weights frozen.

\paragraph{From CTE to Training Weights.}
Let $\mathrm{CTE}_{\ell}$ denote the aggregated, precomputed causal tracing effect of layer $\ell$, obtained by averaging token-level CTE scores over negation-bearing tokens and samples (see Appendix~\ref{app:construction} for details). To ensure stable optimization, we normalize these scores across layers using min--max normalization,
\[
\alpha_{\ell} = 
\frac{\mathrm{CTE}_{\ell} - \min_k \mathrm{CTE}_{k}}
{\max_k \mathrm{CTE}_{k} - \min_k \mathrm{CTE}_{k}},
\]
where $k$ indexes all layers of the text encoder.
This yields layer weights $\alpha_{\ell} \in [0,1]$. Bounding the weights prevents excessively large updates and avoids disrupting the learning dynamics. We retain a global learning rate and treat $\alpha_{\ell}$ as a relative modulation factor rather than a replacement for standard optimization hyperparameters.

\begin{figure*}[t]
    \centering
    \includegraphics[width=0.9\textwidth]{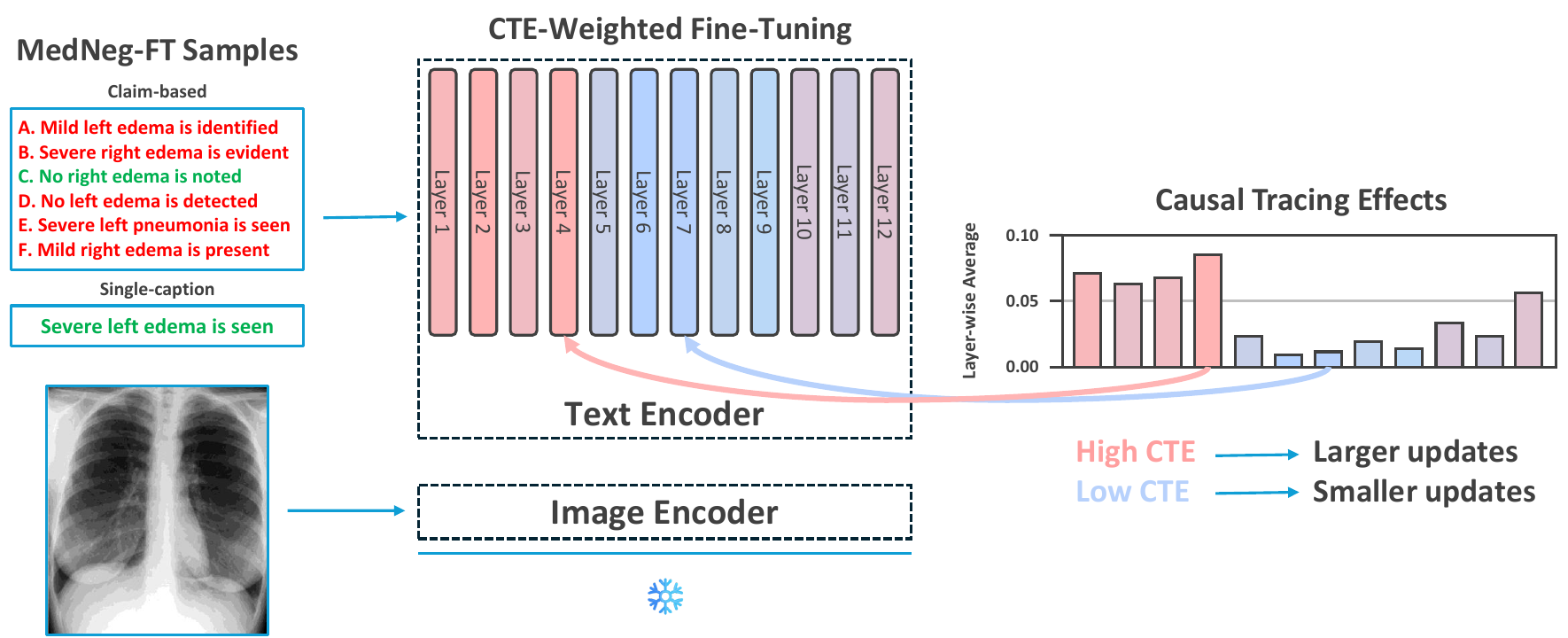}
    \caption{
    \textbf{Overview of Negation-Aware Selective Training (NAST).}
    Layer-wise causal tracing estimates each text encoder layer’s contribution to negation processing. 
    The resulting CTE scores guide fine-tuning on the contextual negation dataset via layer-specific gradient scaling, assigning larger updates to high-CTE layers and smaller updates to low-CTE layers.
    }
    \label{fig:nast_pipeline}
\end{figure*}

\paragraph{CTE-Weighted Gradient Updates.}
Let $g_{\ell}$ denote the gradient of the LoRA parameters at layer $\ell$. During fine-tuning, we apply CTE-aware scaling to obtain
\[
\tilde{g}_{\ell} = \alpha_{\ell}^{\beta} \cdot g_{\ell},
\]
where $\beta > 0$ is a sharpness hyperparameter controlling how strongly updates are concentrated on high-CTE layers relative to low-CTE layers. Larger values of $\beta$ accentuate differences between layers with high and low causal contribution, while smaller values smooth updates across layers. The optimizer then proceeds using $\tilde{g}_{\ell}$ with the original learning rate, effectively inducing layer-specific learning rates grounded in causal interpretability (Figure \ref{fig:nast_pipeline}).

\paragraph{Training Objectives.}
We fine-tune the model using two types of negation-enriched supervision.

\textbf{Single-caption negation supervision.}
For image--caption pairs containing explicit negation, we adopt the standard CLIP contrastive objective. Given a batch
$\mathcal{B}_{\text{cap}} = \{(I_i, T_i)\}_{i=1}^{N}$,
we compute the $i$-th image--text similarity matrix and apply the symmetric contrastive loss
$\mathcal{L}_{\text{CLIP}}(\mathcal{B}_{\text{cap}})$,
encouraging high similarity for matched pairs and low similarity for mismatches.

\textbf{Claim-based caption supervision.}
For samples paired with a set of $K$ mutually exclusive claims
$\{T_{i,1}, \ldots, T_{i,K}\}$ (with exactly one correct claim), we compute cosine similarities between the image $I_i$ and each claim caption as logits $\{\ell_{i,1}, \ldots, \ell_{i,K}\}$. We then apply a claim-ranking loss,
\[
\mathcal{L}_{\text{claim}} =
-\frac{1}{M} \sum_{i=1}^{M}
\log \frac{\exp(\ell_{i,c_i})}
{\sum_{j=1}^{K} \exp(\ell_{i,j})},
\]

where $M$ denotes the number of image--claim sets in the batch and $c_i$ denotes the index of the correct claim for image $i$. This objective encourages the model to assign higher similarity to the semantically correct claim than to closely matched negated or affirmative alternatives.

\paragraph{Combined Objective.}
The final training loss is a weighted combination of the two objectives,
\[
\mathcal{L}_{\text{total}} =
\lambda \, \mathcal{L}_{\text{CLIP}}
+ (1 - \lambda) \, \mathcal{L}_{\text{claim}},
\]
with all gradients modulated by the CTE-based scaling described above. This approach integrates causal interpretability directly into the optimization process, focusing adaptation on layers that are causally responsible for negation understanding while preserving overall multimodal alignment.

\section{Experiments and Results}
\label{sec:results}

We evaluate NAST on contextual negation understanding in medical vision--language models, focusing on both retrieval and claim-based reasoning tasks. 

\subsection{Implementation Details}
\label{res:details}
All experiments are conducted using CLIP ViT-B/32 \citep{radford2021clip} as the base model. Images are resized to $224 \times 224$ and normalized following the standard CLIP preprocessing pipeline. Text inputs are tokenized using the CLIP tokenizer with a maximum sequence length of 77 tokens. Fine-tuning is performed using LoRA-based adapters applied to the attention projection layers of the text encoder, while the image encoder remains frozen. Unless otherwise stated, we use a LoRA rank of $r=8$ and scaling factor $\gamma=16$. This design enables parameter-efficient adaptation while preserving the pretrained backbone.

Optimization is performed using AdamW \citep{loshchilov2017decoupled} with a learning rate of $1 \times 10^{-4}$, weight decay of $10^{-2}$, and batch size of 64. Models are trained for 5 epochs, and the best checkpoint is selected based on validation performance. For NAST, the CTE weighting coefficient $\beta$ is set to $0.5$ (see Appendix \ref{app:further} for sensitivity analysis), and the combined-loss weighting $\lambda$ is set to 0.5. Causal tracing scores are computed once, prior to fine-tuning, on the fixed probing set described in Section~\ref{sec:causal-tracing} and Appendix~\ref{app:construction}. Each experiment is repeated using 3 random seeds (controlling data shuffling, LoRA initialization, and batch sampling), and we report the mean ± standard deviation for all metrics unless otherwise noted. Experiments are conducted on a single NVIDIA RTX 4070 GPU.

\subsection{Experimental Setup}
\label{res:setup}
We compare our approach to CLIP \citep{radford2021clip} and three representative negation-aware variants: NegCLIP \citep{yuksekgonul2022and}, ConCLIP \citep{singh2024learn}, and the negation-focused CLIP model (NegBench) by \citep{alhamoud2025vlmnegation}. All models are initialized from publicly released checkpoints. In addition, we include two baselines to contextualize NAST. First, \emph{CLIP + Negation Prompting}, which uses both affirmative and negated prompts (e.g., ``There is X'' vs.\ ``There is no X'') at inference and selects the higher similarity score. Second, \emph{CLIP + Standard Fine-tuning}, which uses the same training setup as NAST but without CTE-based weighting (claim-based samples excluded, as they are incompatible with the CLIP contrastive loss), isolating its effect. Baselines are evaluated either directly (for pretrained models) or after fine-tuning, where applicable.

\subsection{Evaluation Protocol}
\label{res:eval}

Evaluation is conducted on the test split of the contextual negation benchmark (Section \ref{sec:dataset}) under two settings: (i) single-caption image--text matching and (ii) claim-based caption sets with multiple clinically plausible descriptions per image. All models encode images and text into a shared embedding space, and similarity is computed using cosine similarity over $\ell_2$-normalized embeddings. For models with a learned temperature, we use the default pretrained value without further tuning.

In the single-caption setting, each image is paired with candidate captions, and performance is measured using Recall@1 (R@1) and Recall@5 (R@5). In the claim-based setting, each image has $K$ (typically $K \in [5,7]$) candidate descriptions (one correct and $K-1$ hard negatives), and accuracy is defined as selecting the highest-scoring caption. For the \emph{CLIP + Negation Prompting} baseline, affirmative and negated prompts are evaluated independently, and the higher similarity score is used for prediction. All methods use identical candidate sets and prompts for fair comparison.

\subsection{Main Results on Contextual Negation Tasks}

Table~\ref{tab:main_results} reports the main results on contextual negation tasks. Across both tasks, NAST consistently outperforms all baselines. While prompt-based methods (\emph{CLIP + Negation Prompting}) and standard fine-tuning provide noticeable improvements over the pretrained CLIP model, they remain significantly below specialized negation-aware approaches.

Among prior negation-focused methods, NegBench \citep{alhamoud2025vlmnegation} performs best, followed by ConCLIP \citep{singh2024learn} and NegCLIP \citep{yuksekgonul2022and}. However, NAST achieves the strongest performance overall, indicating that explicitly guiding optimization using causal tracing signals is more effective than architectural or data-level modifications alone. Notably, the relative gains of NAST are larger in the claim-based setting than in retrieval, suggesting that CTE-weighted updates particularly enhance fine-grained polarity discrimination rather than merely improving global image--text alignment.

One might ask whether NAST's gains over the other models partly reflect differences in training data rather than the CTE-weighted optimization itself. We directly control for this in Appendix~\ref{app:further}, where we retrain NegBench on our own MedNeg-FT data under matched conditions; NAST retains a clear advantage even in this controlled setting.

\begin{table}[t]
\centering
\caption{Main results on contextual negation tasks (\%).}
\label{tab:main_results}
\begin{tabular}{lccc}
\toprule
Model & R@1 $\uparrow$ & R@5 $\uparrow$ & Claim Acc. $\uparrow$ \\
\midrule
CLIP & $23.5 \pm 0.5$ & $34.7 \pm 0.6$ & $24.6 \pm 0.6$ \\
CLIP + Negation Prompting & $30.8 \pm 0.5$ & $45.2 \pm 0.4$ & $33.7 \pm 0.7$ \\
CLIP + Fine-tuning & $34.6 \pm 0.6$ & $49.8 \pm 0.7$ & $38.9 \pm 0.7$ \\
NegCLIP & $36.2 \pm 0.6$ & $52.4 \pm 0.5$ & $41.3 \pm 0.7$ \\
ConCLIP & $39.7 \pm 0.5$ & $55.8 \pm 0.7$ & $44.9 \pm 0.6$ \\
NegBench & $43.1 \pm 0.6$  & $59.2 \pm 0.5$ & $48.7 \pm 0.7$ \\
\textbf{NAST (Ours)} & \bm{$49.5 \pm 0.4$} & \bm{$65.7 \pm 0.5$} & \bm{$55.6 \pm 0.4$} \\
\bottomrule
\end{tabular}
\end{table}

\subsection{Performance on Non-Negated Captions}

To verify that improvements in negation sensitivity do not come at the expense of general vision--language alignment, we evaluate all models on the \emph{affirmative-only} subset of the contextual benchmark. This subset contains samples without explicit negation cues.

Table~\ref{tab:non_negated_results} shows that NAST maintains strong performance on non-negated captions. Compared to standard fine-tuning, NAST achieves slightly better or comparable results, indicating that CTE-weighted adaptation does not degrade general alignment. Prompt-based methods and negation-aware baselines also improve over CLIP, but the gains are smaller than those observed with NAST. These findings suggest that NAST selectively improves negation understanding while preserving performance on standard descriptive tasks.

\begin{table}[t]
\centering
\caption{Performance on affirmative-only (non-negated) captions (\%).}
\label{tab:non_negated_results}
\begin{tabular}{lccc}
\toprule
Model & R@1 $\uparrow$ & R@5 $\uparrow$ & Claim Acc. $\uparrow$ \\
\midrule
CLIP & $41.8 \pm 0.6$ & $58.6 \pm 0.6$ & $46.2 \pm 0.7$ \\
CLIP + Negation Prompting & $45.3 \pm 0.6$ & $62.1 \pm 0.7$ & $49.8 \pm 0.5$ \\
CLIP + Fine-tuning & $50.6 \pm 0.5$ & $67.2 \pm 0.7$ & $55.1 \pm 0.6$ \\
NegCLIP & $49.7\pm 0.5$ & $66.3 \pm 0.5$ & $54.1 \pm 0.6$ \\
ConCLIP & $51.2\pm 0.6$ & $68.0 \pm 0.5$ & $55.6 \pm 0.7$ \\
NegBench & $53.8\pm 0.5$ & $70.4 \pm 0.5$ & $58.9 \pm 0.6$ \\
\textbf{NAST (Ours)} & \bm{$54.6 \pm 0.5$} & \bm{$71.2 \pm 0.4$} & \bm{$59.8 \pm 0.5$} \\
\bottomrule
\end{tabular}
\end{table}

\subsection{Reduction of Affirmative--Negation Performance Gap} To directly quantify improvements in polarity sensitivity on an independent evaluation set, we measure the performance gap between affirmative and negated formulations using MedNeg-Bench (Section~\ref{sec:medneg-bench}), which is entirely held out from MedNeg-FT at the patient level. The accuracy is computed separately on the affirmative and negated MCQ sets, and the difference is reported. A smaller gap indicates better robustness to negation. Table~\ref{tab:gap_reduction} reports the resulting gap values. NAST yields the smallest gap, reducing the disparity between affirmative and negated performance. Importantly, this reduction is achieved by improving performance on negated samples rather than sacrificing affirmative accuracy, indicating that gains stem from improved negation understanding rather than global calibration shifts.

\begin{table}[t]
\centering
\caption{Affirmative--Negation gap on MedNeg-Bench (Accuracy \%). Lower is better.}
\label{tab:gap_reduction}
\begin{tabular}{lc}
\toprule
Model & Gap (Affirmative -- Negation) \\
\midrule
CLIP & $15.4 \pm 0.8$ \\
CLIP + Negation Prompting & $11.3 \pm 0.7$ \\
CLIP + Fine-tuning & $9.8 \pm 0.8$ \\
NegCLIP & $8.5 \pm 0.7$ \\
ConCLIP & $7.6 \pm 0.7$ \\
NegBench & $6.3 \pm 0.6$ \\
\textbf{NAST (Ours)} & \bm{$2.8 \pm 0.5$} \\
\bottomrule
\end{tabular}
\end{table}

\subsection{Analysis}
\label{res:analysis}

\paragraph{Generalization Across Contrastive Medical VLMs}

To evaluate whether NAST generalizes beyond a single backbone, we apply it to two additional contrastive medical vision--language models. In addition to the CLIP-based model used in previous experiments, we consider PubMedCLIP \citep{eslami2023pubmedclip} and BioMedCLIP \citep{zhang2023biomedclip}, both pretrained on large-scale biomedical image–text corpora but differing in pretraining data source and scale. We follow the same fine-tuning procedure and evaluation protocol as described in Sections \ref{res:details} and \ref{res:setup}, without modifying any hyperparameters. This ensures that improvements reflect the generality of NAST rather than model-specific tuning.

\begin{table}[h]
\centering
\caption{Performance of NAST across different contrastive medical VLMs (\%).}
\label{tab:cross_model}
\begin{tabular}{l|l|ccc}
\toprule
Model & Method & R@1 $\uparrow$ & R@5 $\uparrow$ & Claim Acc. $\uparrow$ \\
\midrule
\multirow{2}{*}{CLIP} 
& Baseline & $23.5 \pm 0.5$ & $34.7\pm 0.6$ & $24.6\pm 0.6$ \\
& + NAST  & \bm{$49.5\pm 0.4$} & \bm{$65.7\pm 0.5$} & \bm{$55.6\pm 0.4$} \\
\midrule
\multirow{2}{*}{PubMedCLIP} 
& Baseline & $32.1\pm 0.6$ & $46.3\pm 0.7$ & $35.8\pm 0.6$ \\
& + NAST  & \bm{$40.6\pm 0.5$} & \bm{$55.9\pm 0.6$} & \bm{$46.1\pm 0.5$} \\
\midrule
\multirow{2}{*}{BioMedCLIP} 
& Baseline & $41.2\pm 0.6$ & $57.8\pm 0.6$ & $46.5\pm 0.5$ \\
& + NAST  & \bm{$46.8\pm 0.5$} & \bm{$63.9\pm 0.6$} & \bm{$52.7\pm 0.5$} \\
\bottomrule
\end{tabular}
\end{table}

As shown in Table \ref{tab:cross_model}, NAST consistently improves performance across all three backbones. While absolute performance differs due to pretraining data and initialization, the relative gains are noticeable across R@1, R@5, and claim accuracy for both BioMedCLIP and PubMedCLIP, with the largest relative gains observed for CLIP due to its weaker baseline. These results indicate that NAST is not tied to a specific backbone but generalizes across contrastive medical vision--language models. The consistent improvements suggest that layer-wise causal signals for negation are shared across architectures, enabling effective transfer of the proposed optimization strategy.

\paragraph{Zero-Shot Transfer to External Classification Benchmarks}
\label{sec:class}
To assess whether the improved negation sensitivity of NAST translates to downstream clinical tasks, we evaluate zero-shot classification on an external dataset without additional fine-tuning. Based on the CheXpert benchmark \citep{irvin2019chexpert} and standard CLIP-style evaluation, we construct textual prompts of the form ``There is [condition]'' and ``There is no [condition]'' for each label, and compute cosine similarity between image and text embeddings. Predictions are obtained by comparing scores of affirmative and negated prompts.

We report the area under the ROC curve (AUC) averaged across conditions, the macro F1 score, and, to specifically evaluate negation handling, F1 on the subset of samples with absent (negated) conditions. As shown in Table~\ref{tab:external_chexpert}, NAST improves overall zero-shot performance, with larger gains on the negated subset, indicating better recognition of absent findings and suggesting that NAST's benefits extend beyond the constructed benchmark to general clinical prediction in a zero-shot setting.

\begin{table}[h]
\centering
\caption{Zero-shot classification performance on CheXpert. ``Neg. F1'' denotes macro F1 score on samples with absent conditions.}
\label{tab:external_chexpert}
\begin{tabular}{l|ccc}
\toprule
Model & AUC $\uparrow$ & F1 $\uparrow$ & Neg. F1 $\uparrow$ \\
\midrule
CLIP (pretrained) & $0.842 \pm 0.006$ & $0.512 \pm 0.012$ & $0.468 \pm 0.014$ \\
\textbf{NAST (Ours)}       & \bm{$0.861 \pm 0.005$} & \bm{$0.537 \pm 0.011$} & \bm{$0.523 \pm 0.010$} \\
\bottomrule
\end{tabular}
\end{table}

\paragraph{Impact of CTE-Weighted Updates on Learning Dynamics}

We analyze how CTE-weighted adaptation influences optimization across layers. Since NAST scales gradients by estimated causal contribution to negation, we examine the distribution of update magnitudes during fine-tuning by computing the average $\ell_2$ norm of parameter updates per layer under uniform and CTE-weighted training. Uniform fine-tuning distributes updates relatively evenly across layers, whereas CTE-weighted training concentrates updates in layers identified as negation-sensitive, while reducing changes in less relevant layers.

Table~\ref{tab:update_norms} reports the proportion of total update magnitude allocated to the top-$k$ negation-sensitive layers (by CTE score). The results show that CTE-weighted training significantly increases update concentration in these layers, confirming that it reshapes learning dynamics rather than merely re-scaling gradients.

\begin{table}[t]
\centering
\caption{Proportion of total update magnitude in top negation-sensitive layers (\%).}
\label{tab:update_norms}
\begin{tabular}{lcc}
\toprule
Method & Top-3 Layers & Top-5 Layers \\
\midrule
Uniform Fine-tuning & 28.4 & 41.7 \\
\textbf{NAST (CTE-Weighted)} & \textbf{52.6} & \textbf{69.3} \\
\bottomrule
\end{tabular}
\end{table}

\section{Conclusion}

We examined negation understanding in medical vision--language models and identified a systematic affirmative bias under polarity-controlled descriptions. To diagnose this issue, we introduced a radiology-specific benchmark that isolates negation effects. We then proposed Negation-Aware Selective Training (NAST), an interpretability-guided fine-tuning strategy that uses layer-wise causal tracing to modulate gradient updates. NAST improves discrimination between affirmative and negated clinical statements while preserving overall vision--language alignment. These results demonstrate that causal interpretability can guide targeted adaptation in safety-critical multimodal systems.

\section{Limitations}

While NAST demonstrates consistent improvements across CLIP-based VLMs, several limitations remain. First, our method is developed and evaluated exclusively on contrastive (CLIP-style) architectures; extending causal-tracing-guided optimization to generative or autoregressive VLMs is not directly evaluated here. That said, we believe such an extension is readily achievable, as the underlying causal tracing procedure is architecture-agnostic in principle and has already been studied in generative vision--language models such as BLIP~\citep{palit2023towards}, suggesting a natural path for adapting NAST beyond contrastive backbones. Second, MedNeg-Bench and MedNeg-FT are both derived from chest X-ray studies, and we do not empirically verify generalization to other imaging modalities (e.g., pathology, MRI, CT); however, our construction pipeline relies only on structured condition labels and curated polarity mappings, which are not inherently specific to radiology and could plausibly transfer to other domains with modality-appropriate label sources. Finally, our experiments use a single architectural family (contrastive, CLIP-style models) and modest-scale LoRA adaptation on a single GPU; further work is needed to assess how NAST scales to larger backbones, full fine-tuning regimes, and broader deployment settings.

% Third, our causal tracing scores are estimated from a fixed probing set restricted to high-certainty, severe-finding cases, and we do not exhaustively characterize how these scores generalize to more varied or ambiguous clinical phrasing; the close agreement we observe between static and dynamically re-estimated CTE scores (Appendix~\ref{app:further}) is suggestive but not conclusive evidence of robustness. 

\acks{This work was supported in part by NSF awards 2443639 and 2552481, and NIH awards P20GM103446 and U54GM104941.}

\newpage
\bibliography{main}

%%%%%%%%%%%%%%%%%%%%%%%%%%%%%%%%%%%%%%%%%%%%%%%%%%%%%%%%%%%%%%%%%%%%%%%%%%%%%%%
%%%%%%%%%%%%%%%%%%%%%%%%%%%%%%%%%%%%%%%%%%%%%%%%%%%%%%%%%%%%%%%%%%%%%%%%%%%%%%%
% APPENDIX
%%%%%%%%%%%%%%%%%%%%%%%%%%%%%%%%%%%%%%%%%%%%%%%%%%%%%%%%%%%%%%%%%%%%%%%%%%%%%%%
%%%%%%%%%%%%%%%%%%%%%%%%%%%%%%%%%%%%%%%%%%%%%%%%%%%%%%%%%%%%%%%%%%%%%%%%%%%%%%%
\newpage
\appendix
\onecolumn

\section{Affirmative Alternatives for Absent Conditions}
\label{app:affirmative}

To enable controlled comparison between negated and affirmative-equivalent clinical descriptions, we curate a set of affirmative alternative phrases corresponding to the absence (label $=0$) of each CheXpert condition. These phrases are designed to express the same underlying clinical state as a negated finding, while avoiding explicit negation cues (e.g., ``no'', ``without''). All alternatives were developed in consultation with two board-certified radiologists to ensure clinical validity, mutual exclusivity, and consistency with radiology reporting conventions.

For conditions labeled as present (label $=1$), we use the exact condition names provided in the CheXpert annotations when generating MCQ answer choices. Two minor exceptions are made for clarity and clinical naturalness: for \textit{Edema}, we use ``pulmonary edema'' instead of the generic term ``edema'', and for \textit{Pleural Other}, we use ``pleural abnormality'' to better reflect standard radiological phrasing.

Table~\ref{tab:affirmative_mapping} lists the affirmative alternative descriptions used for absent conditions.

\begin{table}[h]
\centering
\caption{Affirmative alternative descriptions used to replace negated clinical findings (label $=0$) in MedNeg-Bench.}
\small
\begin{tabular}{ll}
\toprule
\textbf{Condition (Absent)} & \textbf{Affirmative Alternative Description} \\
\midrule
Atelectasis & well-aerated expanded lungs \\
Cardiomegaly & normal heart size \\
Consolidation & clear lung parenchyma \\
Edema & normal pulmonary vascularity \\
Enlarged Cardiomediastinum & normal mediastinal contours \\
Fracture & intact bony structures \\
Lung Lesion & homogeneous lung parenchyma \\
Lung Opacity & well-aerated lung fields \\
Pleural Effusion & sharp costophrenic angles \\
Pleural Other & smooth pleural surfaces \\
Pneumonia & aerated alveoli \\
Pneumothorax & fully expanded lungs \\
Support Devices & device-free chest \\
\bottomrule
\end{tabular}
\label{tab:affirmative_mapping}
\end{table}

\section{MedNeg-Bench Construction Details}
\label{app:full}

This appendix provides the complete construction procedure for MedNeg-Bench, including raw choice generation, question synthesis, affirmative rewriting, decoding configuration, and validation procedures.

\paragraph{Raw Choice Generation.}
From the filtered CheXpert subset (Section \ref{sec:benchmark}), we randomly select three conditions per study from those annotated as present or absent. The ground-truth binary values of these conditions form the correct answer choice. We then generate three additional contrastive label configurations by randomly permuting the binary values, resulting in four candidate choices per sample. An example set of raw choices is shown below:
\begin{quote}
Lung Lesion $=0$, Cardiomegaly $=0$, Pneumothorax $=1$ \\
Lung Lesion $=0$, Cardiomegaly $=0$, Pneumothorax $=0$ \\
Lung Lesion $=1$, Cardiomegaly $=1$, Pneumothorax $=1$ \\
Lung Lesion $=1$, Cardiomegaly $=0$, Pneumothorax $=1$
\end{quote}
These configurations define the semantic content of the MCQ options and enable controlled comparison between correct and incorrect clinical states.

\paragraph{Negated MCQ Generation.}
The raw choices, together with a condition-name mapping (using standard CheXpert condition names, with minor adjustments for \textit{Edema} and \textit{Pleural Other} as described in Appendix~\ref{app:affirmative}), are provided to GPT-5.1 to generate a radiology-style MCQ with explicit negation.

Generation is performed using deterministic decoding with temperature $T=0.2$ and top-$p=0.95$, with a maximum length of 128 tokens. We use the default system configuration of the API without additional fine-tuning or prompt chaining. The prompt used for generation is shown below:

\begin{quote}
\small
\begin{verbatim}
You are a radiologist creating a multiple-choice question based on chest 
radiograph findings.

Task:
- Write a clear and neutral question stem that asks which option correctly 
  describes the image findings. Keep the question generic, without revealing 
  information about what is or isn't present.
- Rephrase each option into natural medical language and radiology report 
  summary.
- For conditions marked =1, use varied ways to express presence, e.g.
  "with Cardiomegaly", "showing Cardiomegaly", "evidence of Cardiomegaly",
  "Cardiomegaly noted".
- For conditions marked =0, use varied ways to express absence, e.g.
  "without Edema", "no Edema", "Edema absent", "free of Edema", "lacking 
  Edema".
- Use natural conjunctions and negations, with patterns like:
  "A and B without C", "A, B, but not C",
  "C absent while A and B present",
  "No C, with A and B",
  "A noted, no sign of B or C".
- State ONLY the presence or absence of the specified conditions,
  and avoid explanatory phrases or redundant descriptions.
- Output only the question and its options (A–D), with no explanations or 
  extra text.

Condition names to use (USE THESE EXACT TERMS ONLY):
{relevant_map}

Ground truth values:
{raw_choices}
\end{verbatim}
\end{quote}

This yields natural radiology-style questions with diverse but controlled negation patterns.

\paragraph{Affirmative-Equivalent Rewriting.}
To isolate the effect of linguistic polarity, we generate an affirmative-equivalent version of each MCQ. The negated question and options, the raw choices, and the affirmative replacement dictionary for absent conditions (Appendix~\ref{app:affirmative}) are provided to GPT-5.1 with strict instructions to modify only negation phrases.

Rewriting uses deterministic decoding with temperature $T=0.0$ to ensure minimal variation and high fidelity to the original structure. The exact prompt is shown below:

\begin{quote}
\small
\begin{verbatim}
You are rewriting radiology multiple-choice options to replace ONLY the 
negation phrases with affirmative alternatives.

Task:
- Keep the overall sentence structure, word order, and style as close as 
  possible to the original.
- ONLY change phrases that negate a condition (e.g., "no X", "without X",
  "X absent", "free of X", "lacking X", "no evidence of X",
  "no radiographic evidence of X").
- Replace negation phrases with the affirmative alternative provided below.
- Make MINIMAL grammatical adjustments ONLY when necessary.
- For conditions that are present (=1), keep the EXACT original wording 
  unchanged.
- Ensure the result is grammatically correct and sounds natural.
- If a negation phrase includes words like "while", "and", or "with",
  adjust minimally to maintain natural grammar.
- If replacing a negated abnormality (=0) with a normal/benign affirmative 
  finding, remove or replace adversative conjunctions (e.g., "but", "however") 
  when needed.
- If replacing a negation results in repeated "with" constructions,
  replace the later "with" with "and" when it improves fluency.

Output format:
A. [rewritten option A]
B. [rewritten option B]
C. [rewritten option C]
D. [rewritten option D]

Do NOT include the question stem, explanations, or extra text.

Affirmative replacements for absent conditions (=0):
{rewrite_map}

Options to rewrite:
{options_text}

Ground truth (1=present, 0=absent):
{raw_choices}
\end{verbatim}
\end{quote}

This process yields MCQ pairs that are structurally identical except for polarity-related expressions.

\paragraph{Post-processing and Filtering.}
Generated MCQs are post-processed to remove formatting inconsistencies and ensure consistent option structure. We apply rule-based checks to verify that:
(i) each condition mentioned in the raw choices appears exactly once per option,
(ii) negation is correctly aligned with the underlying binary labels, and
(iii) no additional clinical concepts are introduced.
Samples failing these checks are discarded and regenerated. In practice, over 92\% of generated MCQs pass validation on the first attempt.

After filtering, MedNeg-Bench comprises 6,965 negated--affirmative MCQ pairs (13,930 MCQs in total). Patients included in MedNeg-Bench are additionally excluded from MedNeg-FT (Section~\ref{sec:dataset}), ensuring that MedNeg-Bench remains a fully held-out evaluation set with respect to all fine-tuned models reported in this paper.

\paragraph{Human Validation and Error Analysis.}
To assess the quality of the generated benchmark, we conduct a manual evaluation on a random subset of 100 MCQ samples. Each sample is reviewed by a medically trained annotator, who verifies whether (i) the correct option aligns with the underlying label configuration and (ii) the negation in each option is logically consistent.

We observe that $95\%$ of samples are fully consistent with the intended clinical semantics. The remaining cases primarily involve minor linguistic ambiguities or uncommon phrasing patterns rather than incorrect label alignment.

Representative issues include:
\begin{itemize}
    \item \emph{Ambiguous conjunctions}, where coordination (e.g., ``A and B without C'') may require careful parsing;
    \item \emph{Redundant phrasing}, where multiple absence expressions reduce clarity;
    \item \emph{Rare wording variations}, which slightly deviate from standard radiology style.
\end{itemize}

These findings indicate that MedNeg-Bench provides a reliable and controlled evaluation setting, with residual noise limited to surface-level linguistic variation.

\paragraph{Stored Benchmark Fields.}
For each sample, we store the subject ID, study ID, question stem, raw binary label choices, the four negated MCQ options, the four affirmative-equivalent MCQ options, and the index of the correct option. Although CLIP-style models operate directly on image--text pairs and do not require an explicit question, we include the question stem to support evaluation of models that accept question--image inputs, enabling broader applicability of the benchmark.

\section{MedNeg-FT Construction Details}
\label{app:fine}

This appendix provides additional details on the MedNeg-FT construction pipeline, including language model configuration, prompt templates, decoding parameters, validation procedures, and example samples to ensure reproducibility.

\paragraph{Language Model and Decoding Setup.}
All captions are generated using LLaMA~3.1~8B (instruction-tuned). The model is used strictly for surface realization: all semantic content is deterministically derived from CAD annotations \citep{zhang2023cadchest}, and the model is explicitly constrained from introducing new findings, locations, or severity descriptors.

We use deterministic decoding with temperature $T=0.0$ and top-$p=1.0$ (greedy decoding) to ensure consistency across generations. The maximum generation length is set to 64 tokens. No repetition penalty or beam search is applied. All generations are performed in full precision using the HuggingFace Transformers implementation. 

\paragraph{Prompt Templates.}
We employ structured prompts that take filled-in clinical facts as input and request concise radiology-style sentences. An example prompt is shown below:

\begin{quote}
\small
\begin{verbatim}
Instruction:
Generate a concise radiology-style sentence from the structured fact below.
Do not introduce new findings, locations, or severity descriptors.

Fact:
Condition: Pleural effusion
Existence: Absent
Location: Right
Severity: None
\end{verbatim}
\end{quote}

For claim-based contrast sets, multiple facts are provided in a single prompt and the model is instructed to generate one sentence per fact, preserving order and structure.

\paragraph{Post-processing and Normalization.}
Generated captions are lowercased and stripped of extraneous punctuation. Synonym normalization is applied to ensure consistency in condition names and anatomical terms (e.g., ``right lung'' vs.\ ``right side''). This step ensures alignment between structured inputs and textual outputs.

\paragraph{Automatic Validation and Regeneration.}
Generated captions are retained only if they satisfy the following checks:
(i) the specified condition is explicitly mentioned,
(ii) the negated or modified attribute (existence, location, or severity) is correctly expressed, and
(iii) no additional clinical concepts or unsupported attributes are introduced.

Validation is implemented using rule-based pattern matching over the generated text. Captions failing any check are discarded and regenerated with the same prompt until a valid output is obtained or a maximum of three attempts is reached. In practice, over 95\% of generations pass validation on the first attempt.

\paragraph{Example Samples.}

\textbf{Existence Negation (Single Caption):}  
Image: CXR  
Caption: \emph{``There is no pulmonary edema.''}

\textbf{Location Negation (Claim-Based Set):}
\begin{itemize}
    \item True: \emph{``There is consolidation in the right lower lobe.''}
    \item False: \emph{``There is consolidation in the left lower lobe.''}
    \item False: \emph{``There is no consolidation.''}
\end{itemize}

\textbf{Severity Negation (Claim-Based Set):}
\begin{itemize}
    \item True: \emph{``There is a small left pneumothorax.''}
    \item False: \emph{``There is a large left pneumothorax.''}
    \item False: \emph{``There is no pneumothorax.''}
\end{itemize}

\paragraph{Dataset Composition.}
The final dataset contains 988,738 image--text pairs in total, consisting of a mixture of claim-based contrast sets and single negated captions. The dataset is balanced across conditions and negation types (existence, location, severity) to ensure diverse and controlled supervision. Data splits are performed at the patient level to prevent leakage, using an approximate 80/10/10 train/validation/test split.

\paragraph{Human Evaluation Protocol and Error Analysis.}
We perform a manual validation study on 150 randomly sampled image--caption pairs from MedNeg-FT. Each sample is annotated by two independent reviewers with medical training, who are asked to assess: (i) whether the caption correctly reflects the structured clinical fact, and (ii) whether the negation (if present) is logically and clinically valid.

We observe an overall correctness rate of $93.3\%$, with inter-annotator agreement of $\kappa = 0.87$. Disagreements are resolved through discussion.

Error analysis reveals three primary failure modes:
(i) \emph{Attribute omission}, where location or severity modifiers are partially expressed (e.g., missing laterality);
(ii) \emph{Ambiguous phrasing}, where the sentence structure admits multiple interpretations; and
(iii) \emph{Rare lexical variation}, where uncommon synonyms reduce clarity without changing core meaning.

Representative failure cases include:
\begin{itemize}
    \item \emph{``There is no opacity.''} (incorrect: lacks required location specificity)
    \item \emph{``There is mild effusion.''} (incorrect: severity mismatch)
    \item \emph{``No consolidation is seen on the left.''} (ambiguous phrasing)
\end{itemize}

Overall, these results indicate that the generation pipeline produces clinically valid captions with high reliability, while residual errors are limited and largely attributable to surface-level realization rather than incorrect underlying semantics.

\section{Construction of Causal Tracing Inputs}
\label{app:construction}

This appendix describes how we construct inputs for causal tracing analyses used to localize negation-sensitive components in CLIP’s text encoder.

\subsection{Data Selection}

We derive causal tracing inputs from the CAD annotations~\citep{zhang2023cadchest}, which provide comprehensive disease annotations for chest radiographs in the MIMIC-CXR-JPG collection~\citep{johnson2024mimiccxrjpg}. We restrict attention to clinically unambiguous cases to ensure that polarity differences reflect clear diagnostic distinctions. Specifically, we select instances annotated with \texttt{level = severe} and high certainty, with
\[
\texttt{probability\_score} \in \{-3, 3\},
\]
where $-3$ denotes confident absence of a finding and $3$ denotes confident presence. This filtering ensures that polarity differences reflect clear semantic distinctions rather than uncertainty or hedging language.

\subsection{Paired Caption Construction}

For each disease finding, we generate paired textual descriptions that differ only in linguistic polarity:
\begin{itemize}
    \item \textbf{Affirmative}: ``There is severe \textit{[disease]}.''  
    \item \textbf{Negated}: ``There is no \textit{[disease]}.''  
\end{itemize}

The affirmative caption explicitly includes the modifier ``severe'' to approximately match the token length and syntactic complexity with the negated form. This follows the preprocessing protocol of~\citep{quantmeyer2024and} and helps ensure that observed activation differences are attributable to negation rather than trivial length or lexical effects.

We include both cases: (i) cases where negation appears in the caption (\texttt{probability\_score} = $-3$, disease absent), and (ii) cases where negation appears in the foil option (\texttt{probability\_score} = $3$, disease present). This design captures negation processing regardless of whether polarity is expressed in the correct response or in its contrastive alternative.

\subsection{Aggregation Across Negation Contexts}

Because negation may appear either in the caption or in the foil, we merge both subsets to obtain a unified characterization of negation-sensitive layers and positions. Aggregating across these contexts yields a comprehensive view of how negation information is encoded and propagated within the text encoder.

In total, this procedure yields $N = 5,013$ paired instances spanning 13 unique disease findings. These instances form the basis for the causal tracing analyses reported in Section \ref{sec:method} and guide the layer selection used in Negation-Aware Selective Fine-Tuning (NAST).

\section{Additional Experiments and Analysis}
\label{app:further}
This section provides additional analyses to further validate the robustness of NAST beyond the main experimental results. We report a sensitivity analysis of the CTE sharpness hyperparameter $\beta$, an evaluation on an expert-annotated subset of CheXpert to assess robustness to label noise and annotation quality, and a controlled comparison against NegBench trained on matched data to isolate the source of NAST's improvements.

\paragraph{Sensitivity to the CTE Sharpness Hyperparameter}
\label{results:sensitivity}
NAST introduces a sharpness hyperparameter $\beta$ (Section \ref{method:adaptaion}) that controls how strongly gradient updates are concentrated on layers with high causal tracing effects (CTEs). Larger values of $\beta$ emphasize high-CTE layers more aggressively, while smaller values yield a more uniform update distribution. To evaluate the sensitivity of NAST to this hyperparameter, we conduct a controlled ablation study by varying $\beta \in \{0.1, 0.3, 0.5, 0.7, 1.0\}$ while keeping all other training settings fixed. We use the same evaluation protocol as in Section \ref{res:eval}, reporting Recall@1 (R@1), Recall@5 (R@5), and claim-based accuracy.

\begin{table}[h]
\centering
\caption{Sensitivity of NAST to the sharpness hyperparameter $\beta$.}
\label{tab:sens}
\begin{tabular}{c|ccc}
\toprule
$\beta$ & R@1 $\uparrow$ & R@5 $\uparrow$ & Claim Acc. $\uparrow$ \\
\midrule
0.1 & $47.8 \pm 0.6$ & $63.9 \pm 0.7$ & $53.8 \pm 0.6$ \\
0.3 & $48.9 \pm 0.5$ & $65.0 \pm 0.6$ & $54.9 \pm 0.5$ \\
0.5 & \bm{$49.5 \pm 0.4$} & \bm{$65.7 \pm 0.5$} & \bm{$55.6 \pm 0.4$} \\
0.7 & $49.0 \pm 0.5$ & $65.1 \pm 0.6$ & $55.1 \pm 0.5$ \\
1.0 & $48.2 \pm 0.6$ & $64.3 \pm 0.7$ & $54.2 \pm 0.6$ \\
\bottomrule
\end{tabular}
\end{table}

Table \ref{tab:sens} shows that performance is stable across a broad range of $\beta$ values, with only modest variation in all metrics. The best performance is achieved at $\beta = 0.5$, suggesting that moderate sharpening of layer-wise updates provides the most effective balance between focusing on causally important layers and preserving broader representation learning. At lower values (e.g., $\beta = 0.1$), updates are distributed more uniformly across layers, reducing the impact of CTE-based modulation. At higher values (e.g., $\beta = 1.0$), updates become overly concentrated on a small subset of layers, which slightly degrades performance. 

Overall, these results indicate that NAST is not highly sensitive to the choice of $\beta$, and performs robustly across a wide range of settings. In all experiments reported in this paper, we use $\beta = 0.5$, selected based on validation performance.

\paragraph{Evaluation on Expert-Annotated Subset (CheXpert-500)}
\label{res:chexpert500}

To further validate robustness to potential label noise, we evaluate model performance on the CheXpert-500 test set \citep{irvin2019chexpert}, a subset with expert-annotated labels.

We follow the same zero-shot classification setup as described in Section \ref{sec:class}, using prompt pairs of the form ``There is [condition]'' and ``There is no [condition]''. Performance is measured using accuracy and macro F1.

\begin{table}[h]
\centering
\caption{Performance on CheXpert-500 (expert-annotated subset).}
\label{tab:chexpert500}
\begin{tabular}{l|cc}
\toprule
Model & Accuracy (\%) $\uparrow$ & F1 (\%) $\uparrow$ \\
\midrule
CLIP (pretrained) & $58.3 \pm 0.9$ & $54.1 \pm 1.0$ \\
\textbf{NAST (Ours)}       & \bm{$63.7 \pm 0.8$} & \bm{$59.2 \pm 0.9$} \\
\bottomrule
\end{tabular}
\end{table}

NAST improves performance on this expert-annotated subset, with gains of +5.4 in accuracy and +5.1 in F1. These results suggest that the improvements are not driven by noisy labels, but reflect more reliable recognition of clinically meaningful distinctions, including negation.

\paragraph{Baseline Fairness under Matched Training Data}

A natural concern is that NAST's improvements over prior negation-aware methods may partly stem from differences in training data rather than the CTE-weighted optimization strategy itself, since NegCLIP, ConCLIP, and NegBench were originally developed using general-domain negation data. To isolate the contribution of our optimization strategy from that of our dataset, we retrain NegBench directly on MedNeg-FT under matched conditions.

Specifically, NegBench's native loss formulation is designed for MCQ-style supervision with a fixed number of candidate claims. To apply it faithfully, we restrict MedNeg-FT to the subset of claim-based samples containing exactly 4 claims per image, discarding samples with claim sets of other sizes. We train both NegBench and NAST exclusively on this subset, using identical data, splits, and candidate sets for both methods. This removes both the domain-transfer confound (both methods now see the same medical, contextual negation data) and the claim-set-size mismatch as confounds, isolating the effect of NAST's CTE-weighted optimization strategy from any advantage attributable to superior training data.

\begin{table}[h]
\centering
\caption{Baseline fairness under matched training data (\%).}
\label{tab:baseline-fairness}
\begin{tabular}{lccc}
\toprule
Model & R@1 $\uparrow$ & R@5 $\uparrow$ & Claim Acc. $\uparrow$ \\
\midrule
NegBench + our data & $46.2 \pm 0.6$ & $61.8 \pm 0.6$ & $52.1 \pm 0.6$ \\
\textbf{NAST (subset)} & \bm{$48.7 \pm 0.5$} & \bm{$64.5 \pm 0.5$} & \bm{$54.3 \pm 0.5$} \\
\bottomrule
\end{tabular}
\end{table}

Even when NegBench is trained on the same medical, contextual negation data as NAST --- eliminating the domain-transfer confound --- NAST still outperforms it by 2.5 points in R@1, 2.7 points in R@5, and 2.2 points in claim accuracy. This indicates that NAST's advantage does not merely reflect access to better training data, but stems from its interpretability-guided, layer-selective optimization strategy: allocating gradient updates according to causal contribution to negation processing yields a genuine training-dynamics benefit beyond what matched data alone provides.

\end{document}